\documentclass[10pt,twocolumn,letterpaper]{article}

\usepackage{cvpr}
\usepackage{times}
\usepackage{epsfig}
\usepackage{graphicx}
\usepackage{amsmath}
\usepackage{amssymb}

\usepackage{multirow}
\usepackage{booktabs}
\usepackage{subcaption}

\usepackage{enumitem}

\usepackage{xcolor}

\def\rot#1{\rotatebox{90}{#1}}

\usepackage{microtype}
\usepackage[pagebackref=true,breaklinks=true,letterpaper=true,colorlinks,bookmarks=false]{hyperref}

\cvprfinalcopy % *** Uncomment this line for the final submission

\def\cvprPaperID{5569} % *** Enter the CVPR Paper ID here

\makeatletter
\newcommand{\printfnsymbol}[1]{%
        \textsuperscript{\@fnsymbol{#1}}%
}
\makeatother

\ifcvprfinal\fi
\begin{document}

%%%%%%%%% TITLE
\title{StegaStamp: Invisible Hyperlinks in Physical Photographs}

\author{Matthew Tancik\thanks{Authors contributed equally to this work.} \hspace{2cm} Ben Mildenhall\printfnsymbol{1} \hspace{2cm} Ren Ng \hspace*{1cm} \\
University of California, Berkeley\\
{\tt\small \{tancik, bmild, ren\}@berkeley.edu}
}

\maketitle
%\thispagestyle{empty}

%%%%%%%%% ABSTRACT
\begin{abstract}

Printed and digitally displayed photos have the ability to hide imperceptible digital data that can be accessed through internet-connected imaging systems. Another way to think about this is physical photographs that have unique QR codes invisibly embedded within them. This paper presents an architecture, algorithms, and a prototype implementation addressing this vision. Our key technical contribution is StegaStamp, a learned steganographic algorithm to enable robust encoding and decoding of arbitrary hyperlink bitstrings into photos in a manner that approaches perceptual invisibility. StegaStamp comprises a deep neural network that learns an encoding/decoding algorithm robust to image perturbations approximating the space of distortions resulting from real printing and photography.  We demonstrates real-time decoding of hyperlinks in photos from in-the-wild videos that contain variation in lighting, shadows, perspective, occlusion and viewing distance.  Our prototype system robustly retrieves 56 bit hyperlinks after error correction -- sufficient to embed a unique code within every photo on the internet. Code is available at \href{https://github.com/tancik/StegaStamp}{https://github.com/tancik/StegaStamp}.

\end{abstract}

%%%%%%%%% BODY TEXT
\section{Introduction}

\begin{figure*}[t]
\begin{center}
\includegraphics[width=1.\textwidth]{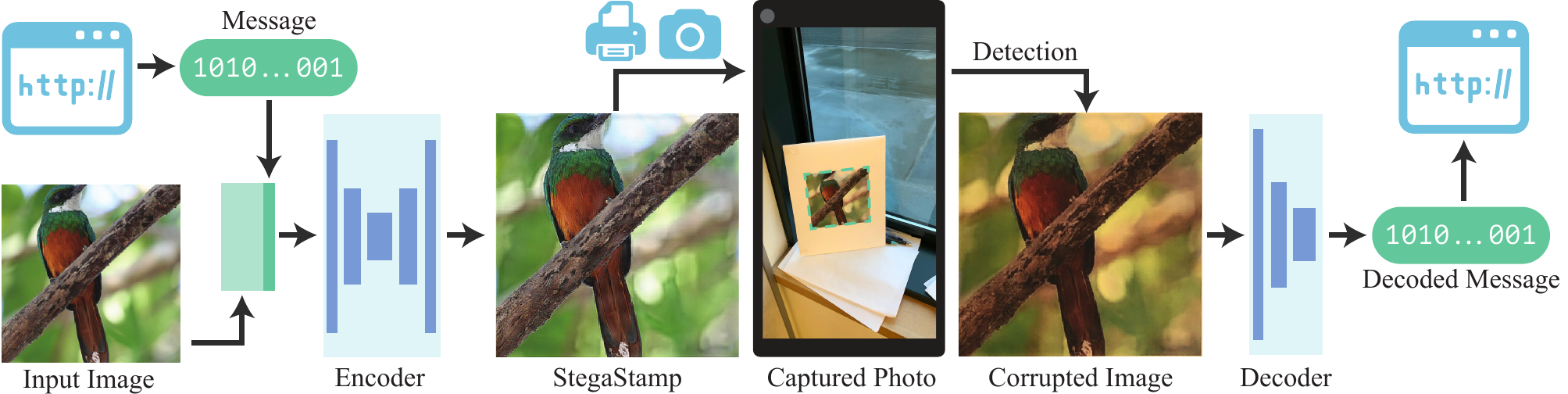}
\end{center}
   \caption{
    Our deep learning system is trained to hide hyperlinks in images. First, an encoder network processes the input image and hyperlink bitstring into a StegaStamp (encoded image). The StegaStamp is then printed and captured by a camera. A detection network localizes and rectifies the StegaStamp before passing it to the decoder network. After the bits are recovered and error corrected, the user can follow the hyperlink. To train the encoder and decoder networks, we simulate the corruptions caused by printing, reimaging, and detecting the StegaStamp with a set of differentiable image augmentations.
    }
\label{fig:pipeline}
\end{figure*}

Our vision is a future in which each photo in the real world invisibly encodes a unique hyperlink to arbitrary information.  This information is accessed by pointing a camera at the photo and using the system described in this paper to decode and follow the hyperlink. In the future, augmented-reality (AR) systems may perform this task continuously, visually overlaying retrieved information alongside each photo in the user's view.
	
Our approach is related to the ubiquitous QR code and similar technologies, which are now commonplace for a wide variety of data-transfer tasks, such as sharing web addresses, purchasing goods, and tracking inventory.  Our approach can be thought of as a complementary solution that avoids visible, ugly barcodes, and enables digital information to be invisibly and ambiently embedded into the ubiquitous imagery of the modern visual world.  

It is worth taking a moment to consider three potential use cases of our system.  First, at the farmer's market, a stand owner may add photos of each type of produce alongside the price, encoded with extra information for customers about the source farm, nutrition information, recipes, and seasonable availability.  Second, in the lobby of a university department, a photo directory of faculty may be augmented by encoding a unique URL for each person's photo that contains the professor's webpage, office hours, location, and directions.  Third, New York City's Times Square is plastered with digital billboards. Each image frame displayed may be encoded with a URL containing further information about the products, company, and promotional deals. 

Figure~\ref{fig:pipeline} presents an overview of our system, which we call StegaStamp, in the context of a typical usage flow.  The inputs are an image and a desired hyperlink.  First, we assign the hyperlink a unique bit string (analogous to the process used by URL-shortening services such as tinyurl.com). Second, we use our StegaStamp encoder to embed the bit string into the target image. This produces an encoded image that is ideally perceptually identical to the input image.  As described in detail in Section~\ref{sec:impl_details}, our encoder is implemented as a deep neural network jointly trained with a second network that implements decoding.  Third, the encoded image is physically printed (or shown on an electronic display) and presented in the real world.  Fourth, a user takes a photo that contains the physical print.  Fifth, the system uses an image detector to identify and crop out all images.  Sixth, each image is processed with the StegaStamp decoder to retrieve the unique bitstring, which is used to follow the hyperlink and retrieve the information associated with the image. 
	
This method of data transmission has a long history in both the steganography and watermarking literatures. We present the first end-to-end trained deep pipeline for this problem that can achieve robust decoding even under ``physical transmission,'' delivering excellent performance sufficient to encode and retrieve arbitrary hyperlinks for an essentially limitless number of images. We extend the traditional learned steganography framework by adding a set of differentiable pixelwise and spatial image corruptions between the encoder and decoder that successfully approximate the space of distortions resulting from ``physical transmission'' (i.e., real printing or display and subsequent image capture).  The result is robust retrieval of 95\% of 100 encoded bits in real-world conditions while preserving excellent perceptual image quality. 
This allows our prototype to uniquely encode hidden hyperlinks for orders of magnitude more images than exist on the internet today (upper bounded by 100 trillion).

\section{Related Work}

\subsection{Steganography}
Steganography is the act of hiding data within other data and has a long history that can be traced back to ancient Greece. Our proposed task is a type of steganography where we hide a code within an image. Various methods have been developed for digital image steganography. Data can be hidden in the least significant bits of the image, subtle color variations, and subtle luminosity variations. Often methods are designed to evade steganalysis, the detection of hidden messages~\cite{fridrich2007statistically, pevny2010using}. We refer the interested reader to surveys~\cite{cheddad2010digital,cox2007digital} that review a wide set of techniques. 

The most relevant work to our proposal are methods that utilize deep learning to both encode and decode a message hidden inside an image~\cite{baluja2017hiding,hayes2017generating,tang2017automatic,wu2018stegnet,zhang2019steganogan,zhu2018hidden,wengrowski2019lfm}. Our method assumes that the image will be corrupted by a display-imaging pipeline between the encoding and decoding steps. With the exception of HiDDeN~\cite{zhu2018hidden} and Light Field Messaging (LFM)~\cite{wengrowski2019light}, small image manipulations or corruptions would render existing techniques useless, as their goal is encoding a large number of bits-per-pixel in the context of perfect digital transmission. HiDDeN introduces various types of noise between encoding and decoding to increase robustness but focuses only on the set of corruptions that would occur through digital image manipulations (e.g., JPEG compression and cropping). For use as a physical barcode, the decoder cannot assume perfect alignment, given the perspective shifts and pixel resampling guaranteed to occur when taking a casual photo. LFM~\cite{wengrowski2019light} obtain robustness using a network trained on a large dataset of manually photographed monitors to undo the camera-display corruptions. Our method does not require this time-intensive dataset capture step and generalizes to printed images, a medium for which collecting training data would be even more difficult.

\subsection{Watermarking}
Watermarking, a form of steganography, has long been considered as a potential way to link a physical image to an Internet resource~\cite{alattar2000}. Early work in the area defined a set of desirable goals for robust watermarking, including invisibility and robustness to image manipulations~\cite{braudaway1997}. Later research demonstrated the significant robustness benefits of encoding the watermark in the log-polar frequency domain~\cite{kang2010,pereira2000robust,pramila2008,zheng2003}. Similar methods have been optimized for use as interactive mobile phone applications~\cite{delgado2013,nakamura2006,pramila2012poster}. Additional work focuses on carefully modeling the printer-camera transform~\cite{pramila2018increasing,solanki2006} or display-camera transform~\cite{fang2019,wengrowski2016calib,yuan2013} for better information transfer. Some approaches to display-camera communication take advantage of the unique properties of this hardware combination such as polarization~\cite{yuan2011}, rolling shutter artifacts~\cite{jo2016}, or high frame rate~\cite{cui2019unseencode}. A related line of work in image forensics explores whether it is possible to use a CNN to detect when an image has been re-imaged~\cite{fan2018}. In contrast to the hand-designed pipelines used in previous work on watermarking, our method automatically \emph{learns} how to hide and transmit data in a way that is robust to many different combinations of printers/displays, cameras, lighting, and viewpoints. We provide a framework for training this system and a rigorous evaluation of its capabilities, demonstrating that it works in many real world scenarios and using ablations to show the relative importance of our training perturbations.

\subsection{Barcodes}
Barcodes are one of the most popular solutions for transmitting a short string of data to a computing device, requiring only simple hardware (a laser reader or camera) and an area for printing or displaying the code. Traditional barcodes are a one dimensional pattern where bars of alternating thickness encode different values. The ubiquity of high quality cellphone cameras has led to the frequent use of two dimensional QR codes to transmit data to and from phones. For example, users can share contact information, pay for goods, track inventory, or retrieve a coupon from an advertisement.

Past research has addressed the issue of robustly decoding existing or new barcode designs using cameras~\cite{liu2008recognition,ohbuchi2004barcode}. Some designs particularly take advantage of the increased capabilities of cameras beyond simple laser scanners in various ways, such as incorporating color into the barcode~\cite{bulan2011color}. Other work has proposed a method that determines where a barcode should be placed on an image and what color should be used to improve machine readability~\cite{myodo2013method}.

Another special type of barcode is specially designed to transmit both a small identifier and a precise six degree-of-freedom orientation for camera localization or calibration, e.g., ArUco markers~\cite{aruco2016,aruco2018}. Hu~\etal~\cite{deepcharuco} train a deep network to localize and identify ArUco markers in challenging real world conditions using data augmentation similarly to our method. However, their focus is robust detection of highly visible preexisting markers, as opposed to robust decoding of messages hidden in arbitrary natural images.

\subsection{Robust Adversarial Image Attacks}
Adversarial image attacks on object classification CNNs are designed to minimally perturb an image in order to produce an incorrect classification. Most relevant to our work are the demonstrations of adversarial examples in the physical world \cite{athalye2017synthesizing, chen2018shapeshifter, evtimov2017robust, jan2019connecting, kurakin2016adversarial, shin2017jpeg, sitawarin2018darts}, where systems are made robust for imaging applications by modeling physically realistic perturbations (i.e., affine image warping, additive noise, and JPEG compression). Jan \etal~\cite{jan2019connecting} take a different approach, explicitly training a neural network to replicate the distortions added by an imaging system and showing that applying the attack to the distorted image increases the success rate. 

These results demonstrate that networks can still be affected by small perturbations after the image has gone through an imaging pipeline. Our proposed task shares some similarities; however, classification targets 1 of $n \approx 2^{10}$ labels, while we aim to uniquely decode 1 of $2^m$ messages, where $m \approx 100$ is the number of encoded bits. Additionally, adversarial attacks typically do not modify the decoder network, whereas we explicitly train our decoder to cooperate with our encoder for maximum information transferal.

\begin{figure}[t]
\begin{center}
\includegraphics[width=\linewidth]{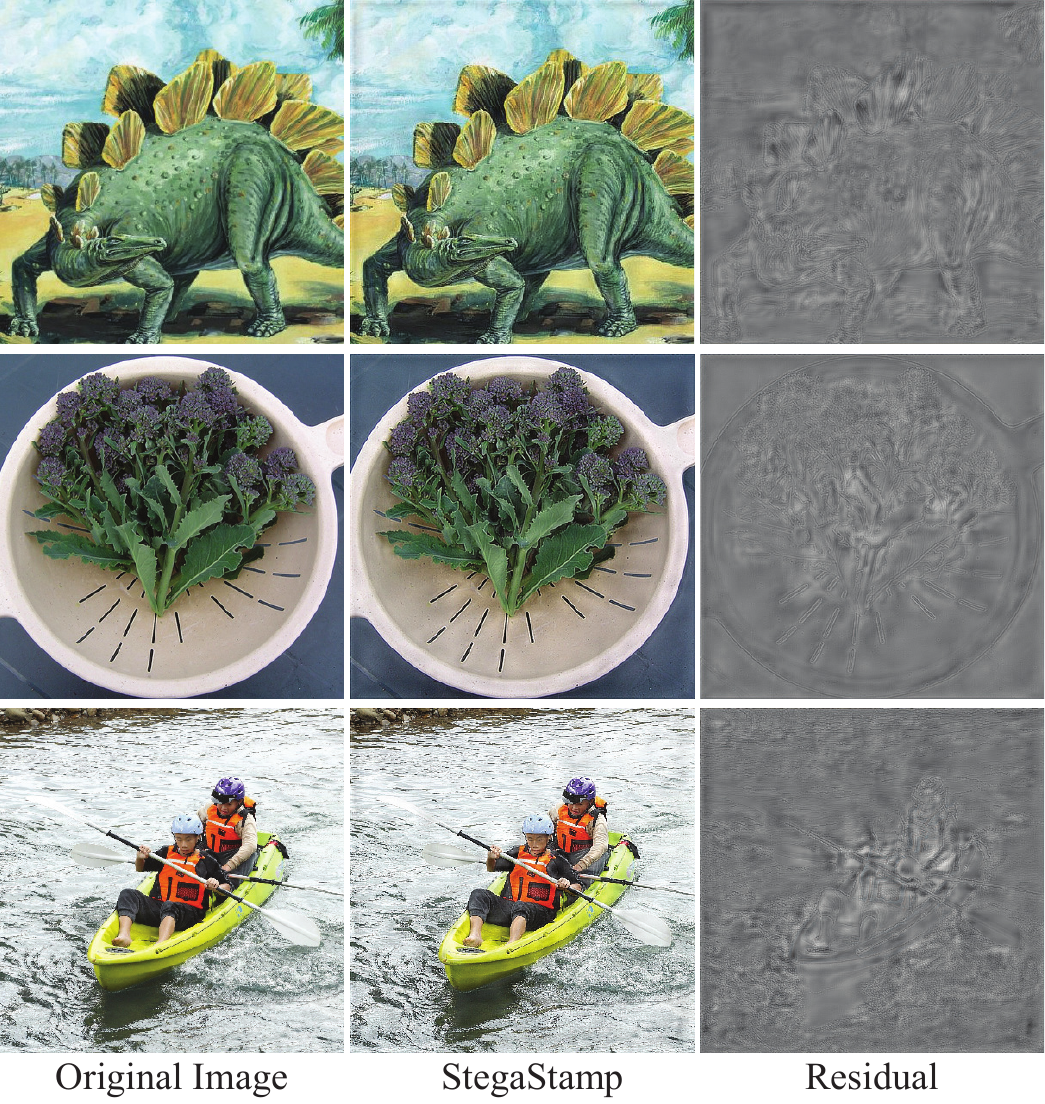}
\end{center}
   \caption{Examples of encoded images. The residual is calculated by the encoder network and added back to the original image to produce the encoded StegaStamp. These examples have 100 bit encoded messages and are robust to the image perturbations that occur through the printing and imaging pipelines. }
\label{fig:examples}
\end{figure}

\begin{figure*}[h!]
\begin{center}
\includegraphics[width=\textwidth]{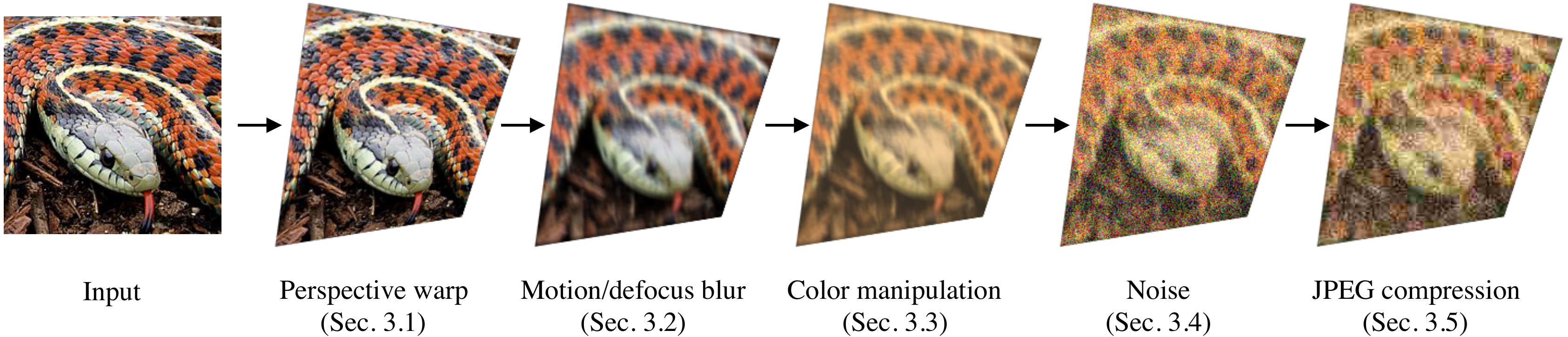}
\end{center}
   \caption{Image perturbation pipeline. During training, we approximate the effects of a physical display-imaging pipeline in order to make our model robust for use in the real world. We take the output of the encoding network and apply the random transformations shown here before passing the image through the decoding network (see Section~\ref{sec:image_modes} for details).}
\label{fig:image_mods}
\end{figure*}

\section{Training for Real World Robustness}
\label{sec:image_modes}

During training, we apply a set of differentiable image perturbations outlined in Figure~\ref{fig:image_mods} between the encoder and decoder to approximate the distortions caused by physically displaying and imaging the StegaStamps. Previous work on synthesizing robust adversarial examples used a similar method to attack classification networks in the wild (termed ``Expectation over Transformation''), though they used a more limited set of transformations~\cite{athalye2017synthesizing}. HiDDeN~\cite{zhu2018hidden} used nonspatial perturbations to augment their steganography pipeline against digital perturbations only. Deep ChArUco~\cite{deepcharuco} used both spatial and nonspatial perturbations to train a robust detector specifically for ChArUco fiducial marker boards. We combine ideas from all of these works, training an encoder and decoder that cooperate to robustly transmit hidden messages through a physical display-imaging pipeline.

\subsection{Perspective Warp}
Assuming a pinhole camera model, any two images of the same planar surface can be related by a homography. We generate a random homography to simulate the effect of a camera that is not precisely aligned with the encoded image marker. To sample a homography, we randomly perturb the four corner locations of the marker uniformly within a fixed range (up to $\pm 40$ pixels, i.e. $\pm 10\%$) then solve for the homography that maps the original corners to their new locations. We bilinearly resample the original image to create the perspective warped image.

\subsection{Motion and Defocus Blur}
Blur can result from both camera motion and inaccurate autofocus. To simulate motion blur, we sample a random angle and generate a straight line blur kernel with a width between 3 and 7 pixels. To simulate misfocus, we use a Gaussian blur kernel with its standard deviation randomly sampled between 1 and 3 pixels.

\subsection{Color Manipulation}

Printers and displays have a limited gamut compared to the full RGB color space. Cameras modify their output using exposure settings, white balance, and a color correction matrix. We approximate these perturbations with a series of random affine color transformations (constant across the whole image) as follows: 
\begin{enumerate}
    \item Hue shift: adding a random color offset to each of the RGB channels sampled uniformly from $[-0.1,0.1]$.  
    \item Desaturation: randomly linearly interpolating between the full RGB image and its grayscale equivalent.
    \item Brightness and contrast: affine histogram rescaling $mx+b$ with $m\sim U[0.5,1.5]$ and $b \sim U[-0.3,0.3]$.
\end{enumerate}
After these transforms, we clip the color channels to $[0,1]$.

\subsection{Noise}
Noise introduced by camera systems is well studied and includes photon noise, dark noise, and shot noise~\cite{Hasinoff2014}. We assume standard non-photon-starved imaging conditions, employing a Gaussian noise model (sampling the standard deviation $\sigma\sim U[0,0.2]$) to account for imaging noise.
% We have found that printing does not introduce significant pixel-wise noise, since the printer will typically act to reduce the noise as the mixing of the ink acts as a low pass filter. 

\subsection{JPEG Compression}

Camera images are usually stored in a lossy format such as JPEG. JPEG compresses images by computing the discrete cosine transform of each $8 \times 8$ block in the image and quantizing the resulting coefficients by rounding to the nearest integer (at varying strengths for different frequencies). This rounding step is not differentiable, so we use the trick from Shin and Song~\cite{shin2017jpeg} for approximating the quantization step near zero with the piecewise function
\begin{equation}
     q(x) = \begin{cases} 
      x^3 &: |x| < 0.5 \\
      x &: |x| \geq 0.5
   \end{cases}
\end{equation}
which has nonzero derivative almost everywhere. We sample the JPEG quality uniformly within $[50,100]$.

\section{Implementation Details}

\label{sec:impl_details}

\subsection{Encoder}
The encoder is trained to embed a message into an image while minimizing perceptual differences between the input and encoded images. We use a U-Net~\cite{unet} style architecture that receives a four channel $400 \times 400$ pixel input (input image RGB channels plus one for the message) and outputs a three channel RGB residual image. The input message is represented as a 100 bit binary string, processed through a fully connected layer to form a $50 \times 50 \times 3$ tensor, then upsampled to produce a $400 \times 400 \times 3$ tensor. We find that applying this preprocessing to the message aids convergence. We present examples of encoded images in Figure~\ref{fig:examples}.

% To enforce minimal perceptual distortion on the encoded StegaStamp, we use an $L_2$ loss, the LPIPS perceptual loss~\cite{zhang2018unreasonable}, and a critic loss calculated between the encoded image and the original image.  

\subsection{Decoder}
The decoder is a network trained to recover the hidden message from the encoded image. A spatial transformer network~\cite{stn} is used to develop robustness against small perspective changes that are introduced while capturing and rectifying the encoded image. The transformed image is fed through a series of convolutional and dense layers and a sigmoid to produce a final output with the same length as the message. The decoder network is supervised using cross entropy loss.

\subsection{Detector}
For real world use, we must detect and rectify StegaStamps within a wide field of view image before decoding them, since the decoder network alone is not designed to handle full detection within a much larger image. We fine-tune an off-the-shelf semantic segmentation network BiSeNet~\cite{BiSeNet} to segment areas of the image that are believed to contain StegaStamps. The network is trained using a dataset of randomly transformed StegaStamps embedded into high resolution images sampled from DIV2K~\cite{div2k}. At test time, we fit a quadrilateral to the convex hull of each of the network's proposed regions, then compute a homography to warp each quadrilateral back to a $400 \times 400$ pixel image for parsing by the decoder.

\subsection{Encoder/Decoder Training Procedure}

\paragraph{Training Data} During training, we use images from the MIRFLICKR dataset~\cite{mirflickr} (resampled to $400\times 400$ resolution) combined with randomly sampled binary messages.

\paragraph{Critic} As part of our total loss, we use a critic network that predicts whether a message is encoded in a image and is used as a perceptual loss for the encoder/decoder pipeline. The network is composed of a series of convolutional layers followed by max pooling. To train the critic, an input image and an encoded image are classified and the Wasserstein loss~\cite{wgan} is used as a supervisory signal. Training of the critic is interleaved with the training of the encoder/decoder. 

\paragraph{Losses}
To enforce minimal perceptual distortion on the encoded StegaStamp, we use an $L_2$ residual regularization $L_R$, the LPIPS perceptual loss~\cite{zhang2018unreasonable} $L_P$, and a critic loss $L_C$ calculated between the encoded image and the original image. We use cross entropy loss $L_M$ for the message. The training loss is the weighted sum of these loss components.
% The training loss is the weighted sum of three image loss terms (residual regularization $L_R$, perceptual loss $L_P$, critic loss $L_C$) and the cross entropy message loss $L_M$:
\begin{equation}
    L=\lambda_R L_R+\lambda_P L_P+ \lambda_C L_C+ \lambda_M L_M
\end{equation}
We find three loss function adjustments to particularly aid in convergence when training the networks:
\begin{enumerate}    \item These image loss weights $\lambda_{R,P,C}$ must initially be set to zero while the decoder trains to high accuracy, after which $\lambda_{R,P,C}$ are increased linearly.
    \item The image perturbation strengths must also start at zero. The perspective warping is the most sensitive perturbation and is increased at the slowest rate.
    \item The model learns to add distracting patterns at the edge of the image (perhaps to assist in localization). We mitigate this effect by increasing the weight of the $L_2$ loss at the edges with a cosine dropoff. 
\end{enumerate}

\begin{figure}[t]
\begin{center}
\includegraphics[width=\linewidth]{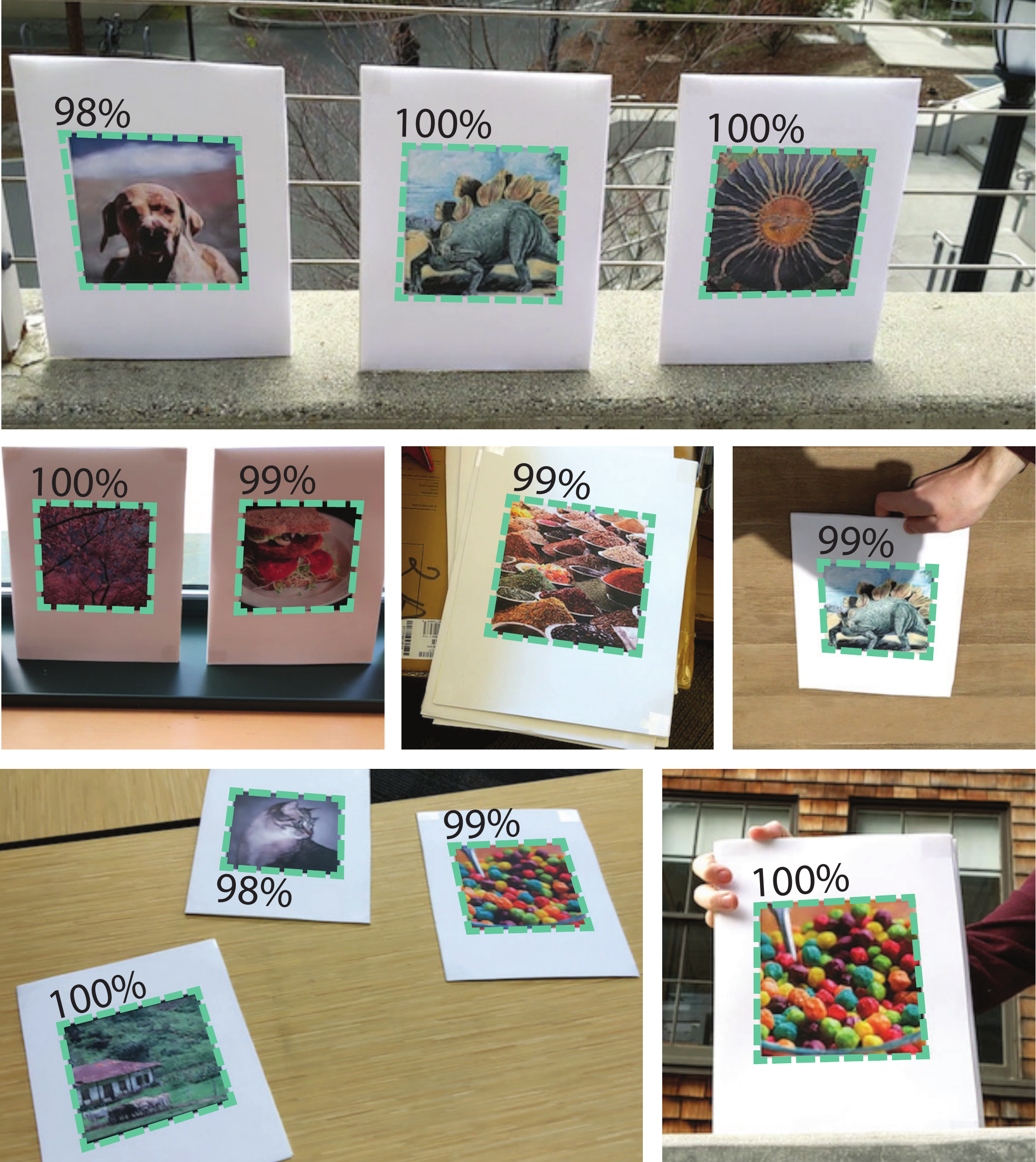}
\end{center}
   \caption{Examples of our system deployed in-the-wild. We outline the StegaStamps detected and decoded by our system and the show message recovery accuracies. Our method works in the real world, exhibiting robustness to changing camera orientation, lighting, shadows, etc. You can find these examples and more in our supplemental video.}
\label{fig:detection}
\end{figure}

%\section{Real and Synthetic Results}
\section{Real-World \& Simulation-Based Evaluation}
\label{sec:results}

We test our system in both real-world conditions and synthetic approximations of display-imaging pipelines. We show that our system works in-the-wild, recovering messages in uncontrolled indoor and outdoor environments. We evaluate our system in a controlled real world setting with 18 combinations of 6 different displays/printers and 3 different cameras. Across all settings combined (1890 captured images), we achieve a mean bit-accuracy of $98.7\%$. We conduct real and synthetic ablation studies with four different trained models to verify that our system is robust to each of the perturbations we apply during training and that omitting these augmentations significantly decreases performance.

\begin{figure}[t]
\begin{center}
\includegraphics[width=\linewidth]{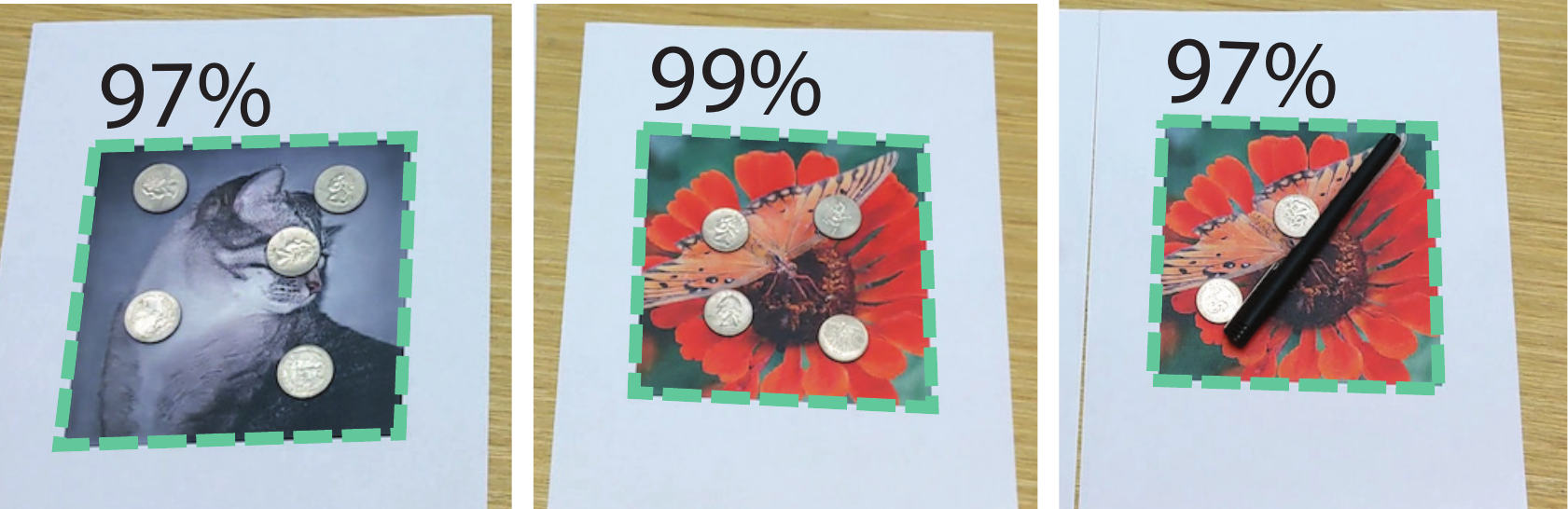}
\end{center}
   \caption{Despite not explicitly training the method to be robust to occlusion, we find that our decoder can handle partial erasures gracefully, maintaining high accuracy.}
\label{fig:occlusion}
\end{figure}

\subsection{In-the-Wild Robustness}
\label{sec:wild}
Our method is tested on handheld cellphone camera videos captured in a variety of real-world environments. The StegaStamps are printed on a consumer printer. Examples of the captured frames with detected quadrilaterals and decoding accuracy are shown in Figure~\ref{fig:detection}. We also demonstrate a surprising level of robustness when portions of the StegaStamp are covered by other objects (Figure~\ref{fig:occlusion}). Please see our supplemental video for extensive examples of real world StegaStamp decoding, including examples of perfectly recovering 56 bit messages using BCH error correcting codes~\cite{bch}. We generally find that if the bounding rectangle is accurately located, decoding accuracy is high. However, it is possible for the detector to miss the StegaStamp on a subset of video frames. In practice this is not an issue, because the code only needs to be recovered once. We expect future extensions that incorporate temporal information and custom detection networks can further improve the detection consistency.

\begin{table}[t]
\centering
\begin{tabular}{ll|l|cccc}
  \multicolumn{3}{l|}{}   & 5th & 25th & 50th & Mean \\ 
\cline{2-7} \multirow{6}{*}{\rotatebox[origin=c]{90}{~Webcam}} & 
\multirow{3}{*}{\rotatebox[origin=c]{90}{~Printer}} & 
Enterprise  & 88\%  & 94\%  & 98\%  & 95.9\% \\ 
 &  & Consumer    & 90\%  & 98\%  & 99\%  & 98.1\% \\ 
 &  & Pro         & 97\%  & 99\%  & 100\%  & 99.2\% \\ 
 & \multirow{3}{*}{\rotatebox[origin=c]{90}{~Screen}} & 
Monitor     & 94\%  & 98\%  & 99\%  & 98.5\% \\ 
 &  & Laptop      & 97\%  & 99\%  & 100\%  & 99.1\% \\ 
 &  & Cellphone   & 91\%  & 98\%  & 99\%  & 97.7\% \\ 
\cline{2-7} \multirow{6}{*}{\rotatebox[origin=c]{90}{~Cellphone}} & 
\multirow{3}{*}{\rotatebox[origin=c]{90}{~Printer}} & 
Enterprise  & 88\%  & 96\%  & 98\%  & 96.8\% \\ 
 &  & Consumer    & 95\%  & 99\%  & 100\%  & 99.0\% \\ 
 &  & Pro         & 97\%  & 99\%  & 100\%  & 99.3\% \\ 
 & \multirow{3}{*}{\rotatebox[origin=c]{90}{~Screen}} & 
Monitor     & 98\%  & 99\%  & 100\%  & 99.4\% \\ 
 &  & Laptop      & 98\%  & 99\%  & 100\%  & 99.7\% \\ 
 &  & Cellphone   & 96\%  & 99\%  & 100\%  & 99.2\% \\ 
\cline{2-7} \multirow{6}{*}{\rotatebox[origin=c]{90}{~DSLR}} & 
\multirow{3}{*}{\rotatebox[origin=c]{90}{~Printer}} & 
Enterprise  & 86\%  & 96\%  & 99\%  & 97.0\% \\ 
 &  & Consumer    & 97\%  & 99\%  & 100\%  & 99.3\% \\ 
 &  & Pro         & 98\%  & 99\%  & 100\%  & 99.5\% \\ 
 & \multirow{3}{*}{\rotatebox[origin=c]{90}{~Screen}} & 
Monitor     & 99\%  & 100\%  & 100\%  & 99.8\% \\ 
 &  & Laptop      & 99\%  & 100\%  & 100\%  & 99.8\% \\ 
 &  & Cellphone   & 99\%  & 100\%  & 100\%  & 99.8\% \\ 
\end{tabular}
\caption{Real world decoding accuracy (percentage of bits correctly recovered) tested using a combination of six display methods (three printers and three screens) and three cameras. We show the 5\textsuperscript{th}, 25\textsuperscript{th}, and 50\textsuperscript{th} percentiles and mean taken over 105 images chosen randomly from ImageNet~\cite{imagenet} with randomly sampled 100 bit messages.}
\label{tab:real_modalities}
\end{table}

\begin{figure}[t]
\begin{center}
\includegraphics[width=\linewidth]{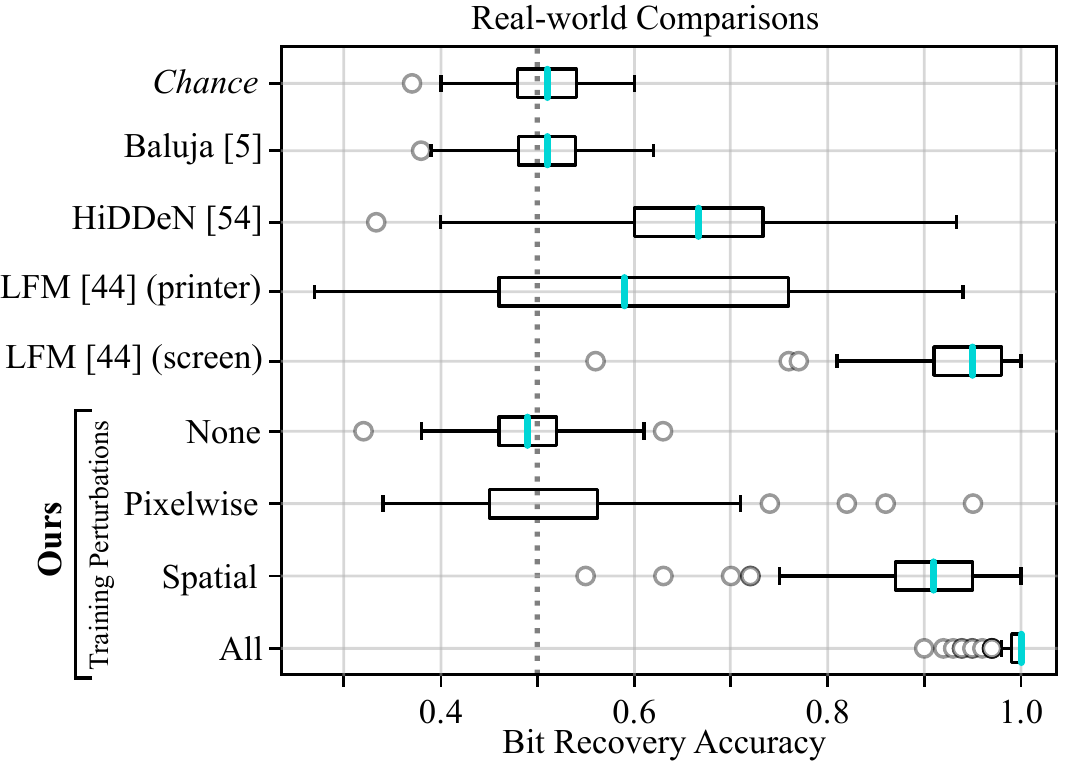}
\end{center}
   \caption{Real world comparisons of variants of our method described in Section~\ref{sec:synth_tests} and competing methods, using the cellphone camera + consumer printer pipeline from Table~\ref{tab:real_modalities}. We show the distribution of random guessing (with its mean of $0.5$ indicated by the dotted line) to demonstrate that the no-perturbations ablation and Baluja~\cite{baluja2017hiding} perform no better than chance. HiDDeN~\cite{zhu2018hidden} uses pixelwise perturbations along with random masking. Adding spatial perturbations is critical for achieving high real-world performance. LFM~\cite{wengrowski2019lfm} works well on screens but fails to generalize to printed media.}
\label{fig:ablations_boxplot}
\end{figure}

\begin{figure*}[h]
\begin{center}
   \centering
   
\includegraphics[width=\textwidth]{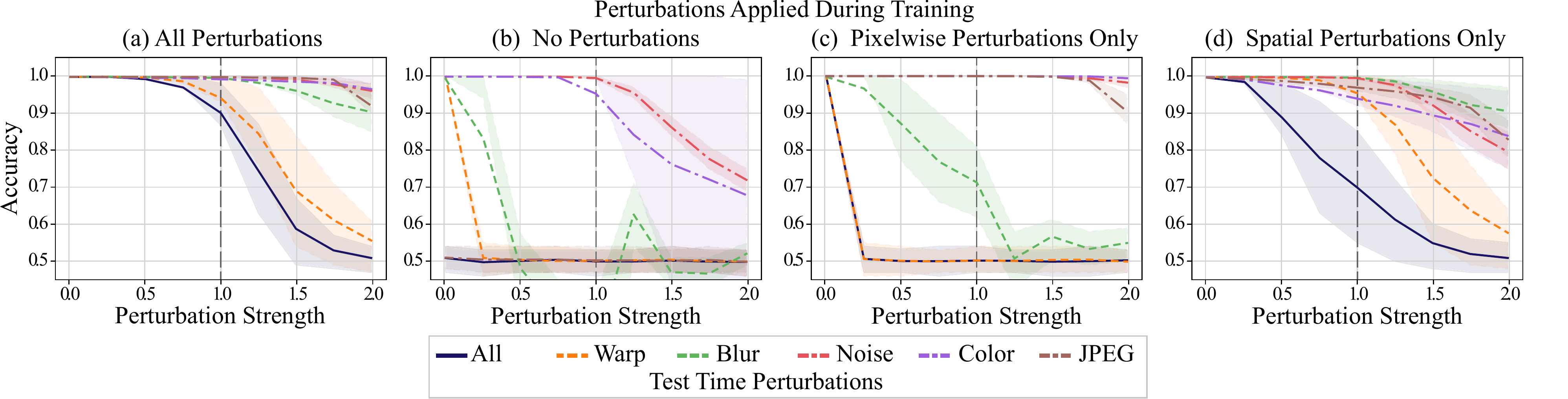}

\end{center}
   \caption{Synthetic ablation tests showing the effect of training with various image perturbation combinations on bit recovery robustness. ``Pixelwise'' perturbations (c) are noise, color transforms, and JPEG compression, and ``spatial'' perturbations (d) are perspective warp and blur. To test robustness across a range of possible degradation, we parameterize the strength of each perturbation on a scale from 0 (weakest) to 1 (maximum value seen during training) to 2 (strongest). Models not trained against spatial perturbations (b-c) are highly susceptible to warp and blur, and the model trained only on spatial perturbations (d) is sensitive to color transformations. The lines show the mean accuracies and the shaded regions shows the 25\textsuperscript{th}-75\textsuperscript{th} percentile range over 100 random images and messages. See Section~\ref{sec:synth_tests} for details.}
\label{fig:synth_plots}
\end{figure*}

\subsection{Controlled Real World Experiments}
\label{sec:real_exp}

In order to demonstrate that our model generalizes from synthetic perturbations to real physical display-imaging pipelines, we conduct a series of test where encoded images are printed or displayed, recaptured by a camera, then decoded. We randomly select 100 unique images from the ImageNet dataset~\cite{imagenet} (disjoint from our training set) and embed random 100 bit messages within each image.
We generate 5 additional StegaStamps with the same source image but different messages for a total of 105 test images.
We conduct the experiments in a darkroom with fixed lighting. The printed images are fixed in a rig for consistency and captured by a tripod-mounted camera. The resulting photographs are cropped by hand, rectified, and passed through the decoder.

The images are printed using a consumer printer (HP LaserJet Pro M281fdw), an enterprise printer (HP LaserJet Enterprise CP4025), and a commercial printer (Xerox 700i Digital Color Press). The images are also digitally displayed on a matte 1080p monitor (Dell ST2410), a glossy high DPI laptop screen (Macbook Pro 15 inch), and an OLED cellphone screen (iPhone X). To image the StegaStamps, we use an HD webcam (Logitech C920), a cellphone camera (Google Pixel 3), and a DSLR camera (Canon 5D Mark II). All devices use their factory calibration settings. Each of the 105 images were captured across all 18 combinations of the 6 media and 3 cameras. The results are reported in Table~\ref{tab:real_modalities}. Our method is highly robust across a variety of different combinations of display/printer and camera; two-thirds of these scenarios yield a median accuracy of 100\% and a 5\textsuperscript{th} percentile accuracy of at least 95\% perfect decoding. Our mean accuracy over all 1890 captured images is $98.7\%$.

Using a test set comprised of the cellphone camera + consumer printer combination, we compare variants of our method (described further in Section~\ref{sec:synth_tests}) to Baluja~\cite{baluja2017hiding}, HiDDeN~\cite{zhu2018hidden}, and LFM~\cite{wengrowski2019lfm} in Figure~\ref{fig:ablations_boxplot}. The variants of our model use the same architecture but are trained with different augmentations; the names \textit{None, Pixelwise, Spatial}, and \textit{All} indicate which categories of peturbations were applied during training. We see that Baluja~\cite{baluja2017hiding}, trained with a minimal amount of augmented noise (similar to our \textit{None} variant) performs no better than guessing. 
HiDDeN~\cite{zhu2018hidden} incorporates augmentations into their training pipeline to increase robustness to perturbations. Their method is trained with a set of pixelwise perturbations along with a ``cropping'' augmentation that masks out a random image region. However, it lacks augmentations that spatially resample the image, and we find that its accuracy falls between our \textit{Pixelwise} and \textit{Spatial} variants.
LFM~\cite{wengrowski2019lfm} specifically trains a ``distortion'' network to mimic the effect of displaying and recapturing an encoded image, trained on a dataset they collect of over 1 million images from 25 display/camera pairs. In this domain (``screen''), we find LFM performs fairly well. However, it does not generalize to printer/camera pipelines (``printer'').
Please refer to the supplement for testing details regarding the compared methods. Among our own ablated variants, we see that training with spatial perturbations alone yields significantly higher performance than only using pixelwise perturbations; however, \textit{Spatial} still does not reliably recover enough data for practical use. Our presented method (\textit{All}), combining both pixelwise and spatial perturbations, achieves the most precise and accurate results by a large margin.

\begin{figure*}[t]
\begin{center}
\includegraphics[width=\textwidth]{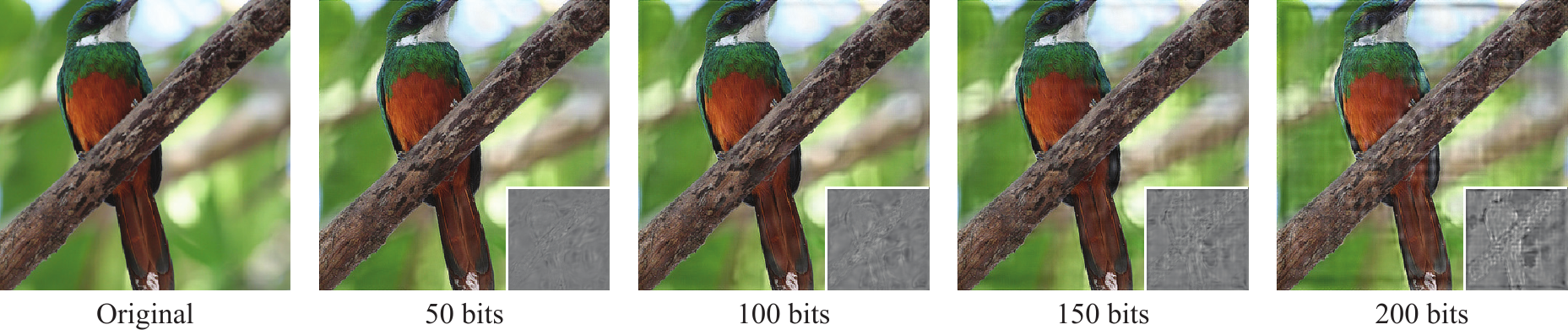}
\end{center}
   \caption{Four models trained to encode messages of different lengths. The inset shows the residual relative to the original image. The perceptual quality decreases as more bits are encoded. We find that a message length of 100 bits provides good image quality and is sufficient to encode a virtually unlimited number of distinct hyperlinks using error correcting codes.}
\label{fig:bit_count_comparision}
\end{figure*}

\subsection{Synthetic Ablation Test}
\label{sec:synth_tests}

We test how training with different subsets of the image perturbations from Section~\ref{sec:image_modes} impacts decoding accuracy in a synthetic experiment (Figure~\ref{fig:synth_plots}). 
We evaluate both our base model (trained with all perturbations) and three additional models (trained with no perturbations, only pixelwise perturbations, and only spatial perturbations). Most work on learned image steganography focuses on hiding as much information as possible, assuming that no corruption will occur prior to decoding (as in our ``no perturbations'' model).

We run a more exhaustive synthetic ablation study over 1000 images to separately test the effects of each training-time perturbation on accuracy.
The results shown in Figure~\ref{fig:synth_plots} follow a similar pattern to the real world comparison test.
The model trained with no perturbations is surprisingly robust to color warps and noise but immediately fails in the presence of warp, blur, or any level of JPEG compression. Training with only pixelwise perturbations yields high robustness to those augmentations but still leaves the network vulnerable to any amount of pixel resampling from warping or blur. On the other hand, training with only spatial perturbations also confers increased robustness against JPEG compression (perhaps because it has a similar low-pass filtering effect to blurring). Again, training with both spatial and pixelwise augmentations yields the best result. 

\begin{table}[t]
\centering
\begin{tabular}{l|cccc}
 & \multicolumn{4}{c}{Message length}   \\
Metric & 50     & 100    & 150    & 200   \\
\hline
PSNR $\uparrow$ & 29.88  & 28.50  & 26.47  & 21.79 \\
SSIM $\uparrow$ & 0.930  & 0.905  & 0.876  & 0.793 \\
LPIPS $\downarrow$ & 0.100  & 0.101  & 0.128  & 0.184 \\
\end{tabular}
\caption{Image quality for models trained with different message lengths, averaged over 500 images. For PSNR and SSIM, higher is better. LPIPS~\cite{zhang2018unreasonable} is a learned perceptual similarity metric, lower is better.}
\label{tab:image_quality}
\end{table}

\subsection{Practical Message Length}
Our model can be trained to store different numbers of bits. In all previous examples, we use a message length of 100. Figure~\ref{fig:bit_count_comparision} compares encoded images from four separately trained models with different message lengths. Larger message are more difficult to encode and decode; as a result, there is a trade off between recovery accuracy and perceptual similarity. The associated image metrics are reported in Table~\ref{tab:image_quality}. When training, the image and message losses are tuned such that the bit accuracy converges to at least 95\%.

We settle on a message length of 100 bits as it provides a good compromise between image quality and information transfer. 
Given an estimate of at least 95\% recovery accuracy, we can encode at least 56 error corrected bits using BCH codes~\cite{bch}. As discussed in the introduction, this gives us the ability to uniquely map every recorded image in history to a corresponding StegaStamp. Accounting for error correcting, using only 50 total message bits would drastically reduce the number of possible encoded hyperlinks to under one billion. The image degradation caused by encoding 150 or 200 bits is much more perceptible.

\subsection{Limitations}

Though our system works with a high rate of success in the real world, it is still many steps from enabling broad deployment. Despite often being very subtle in high frequency textures, the residual added by the encoder network is sometimes perceptible in large low frequency regions of the image. Future work could improve upon our architecture and loss functions to generate more subtle encodings.

Additionally, we find our off-the-shelf detection network to be the bottleneck in our decoding performance during real world testing. A custom detection architecture optimized end to end with the encoder/decoder could increase detection performance. The current framework also assumes that the StegaStamps will be single, square images for the purpose of detection. We imagine that embedding multiple codes seamlessly into a single, larger image (such as a poster or billboard) could provide even more flexibility.

%-----------------------------------------------------------------------------------
\section{Conclusion}
We have presented an end-to-end deep learning framework for encoding 56 bit error corrected hyperlinks into arbitrary natural images. Our networks are trained through an image perturbation module that allows them to generalize to real world display-imaging pipelines. We demonstrate robust decoding performance on a variety of printer, screen, and camera combinations in an experimental setting. We also show that our method is stable enough to be deployed in-the-wild as a replacement for existing barcodes that is less intrusive and more aesthetically pleasing.

\section{Acknowledgments}
We thank Coline Devin and Cecilia Zhang for acting in our supplemental video and Utkarsh Singhal and Pratul Srinivasan for useful feedback. BM is supported by a Hertz Fellowship and MT is supported by NSF GRFP.

{\small
\bibliographystyle{ieee_fullname}
\bibliography{egbib}
}

\clearpage

\appendix

\section{StegaStamp Examples}

\begin{figure*}[h]
\begin{center}
\includegraphics[width=.90\linewidth]{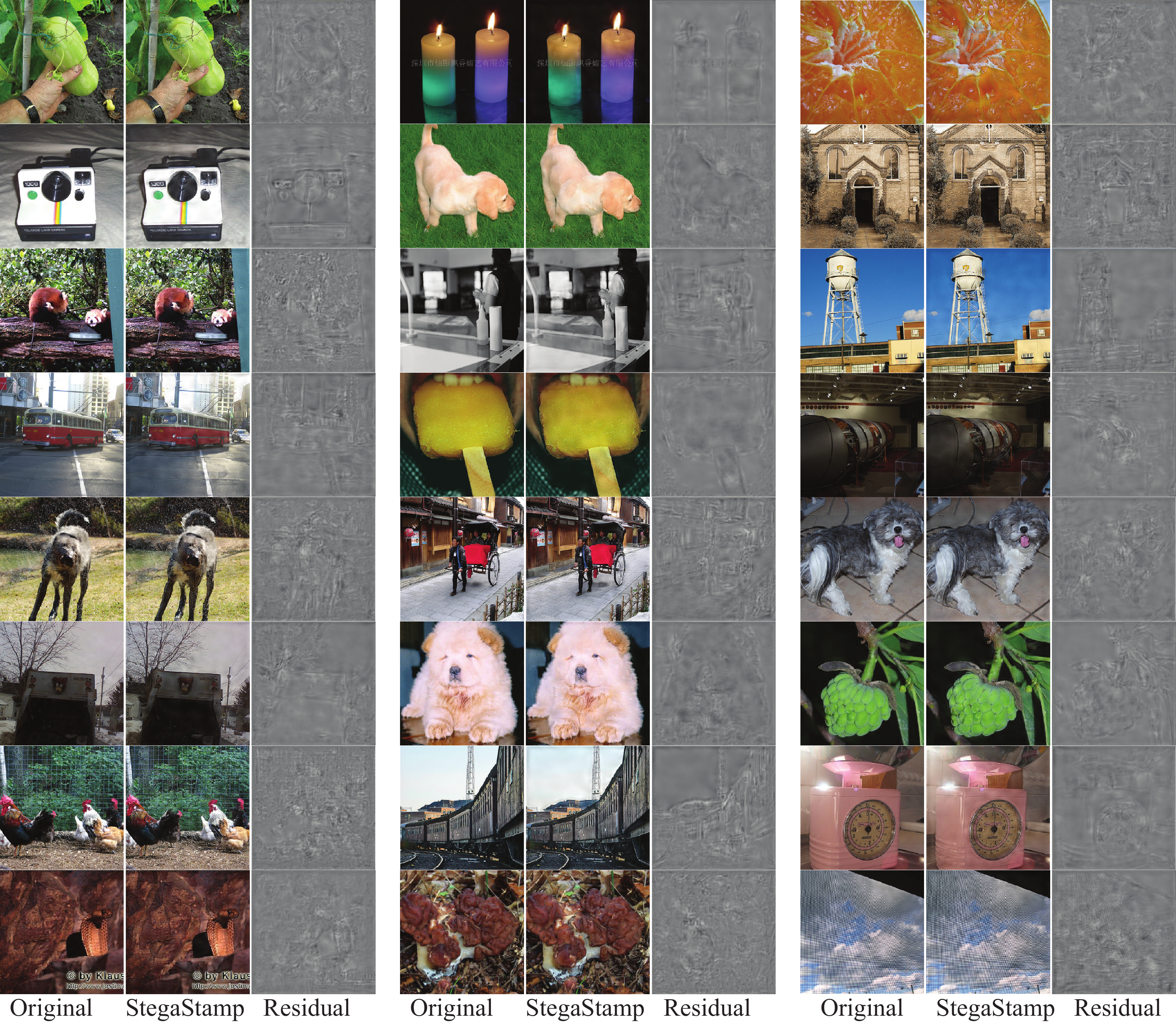}
\end{center}
  \caption{Additional examples of encoded images and their residuals.}
\label{fig:examples_1}
\end{figure*}

See Figure~\ref{fig:examples_1} for additional examples of encoded images and their residuals.

\section{Supplemental Videos}
\noindent
\href{https://youtu.be/E8OqgNDBGO0}{https://youtu.be/E8OqgNDBGO0}

This video provides an overview of StegaStamps with example use cases and a condensed demonstration of in-the-wild results.
\\\\
\noindent
\href{https://youtu.be/jpbRhOH3D9Y}{https://youtu.be/jpbRhOH3D9Y}

This video is a compilation of multiple in-the-wild captures. The first set of clips visualizes the output bounding polygons along with the percentage of bits recovered correctly out of 100. We filter the output to only show detections where the bit accuracy is greater than 70 percent. We note that the messages are regularly recovered with greater than 90\% accuracy when they are properly detected. The second set of clips demonstrates the used of BCH error correction~\cite{bch} to robustly detect and correct recovered codes. The transmitted data consists of 56 message bits and 40 error correcting bits. When the accuracy is greater than 95\% (fewer than 5 corrupted bits), the original 56-bit message can be recovered exactly. If too many bits are corrupted, the error correcting fails and we filter out the proposal. The video represents successfully decoded StegaStamps with green polygons. The decoded code is printed above the polygon. Note that for most real world applications, it is only necessary to recover the code in a single video frame to count it as successfully scanned.

\section{Comparison Details}
We compare our method to Baluja~\cite{baluja2017hiding}, HiDDeN~\cite{zhu2018hidden}, and LFM~\cite{wengrowski2019lfm}. Baluja was designed to hide images within images, which differs from our task of hiding a bitstring within an image. To account for this, we convert our 100 bit message into a $10 \times 10$ grid of ones and zeros that is upscaled to the resolution of the cover image. During decoding we round the model output to 0 and 1 and take the mode within each upscaled block. As the original model was trained to hide natural images, we retrain the model from scratch to hide our bitstring grids.

HiDDeN was trained to hide 30 bit messages in $128 \times 128$ pixel images. We observed a significant drop in accuracy when we trained a model to hide 100 bit messages in $400 \time 400$ pixel images, therefore we report accuracy results on the 30 bit in $128^2$ image version.

LFM~\cite{wengrowski2019lfm} was trained to encode 1024 bit messages as $4 \times 4$ pixel blocks in a $256 \times 256$ pixel image. To encode our 100 bit message, we allocated 9 blocks for each message bit (we therefore only use a $244 \times 244$ pixel subset of the image). We average and round the 9 block predictions to recover the message bit.

Each compared method encodes a different length message into a different size image. However, if we treat the mean bit recovery accuracy (first column in Table~\ref{tab:bpp}) as the crossover probability $p$ in a binary symmetric channel, we can use information theory to calculate the channel capacity (with unit ``bits''):
\begin{equation}
    C(p) = 1 - \left(-p\log_2{p} - (1-p)\log_2{1-p}\right)
\end{equation}
If we divide $C(p)$ by the number of pixels $N_{pix}$ in the original image, we get the expected number of bits-per-pixel transmitted by that method. Multiplying $\frac{C(p)}{N_{pix}}$ by $10^6$ yields our bits-per-megapixel metric in the second column of Table~\ref{tab:bpp}.

\begin{table}[t]
  \newcommand{\multirot}[1]{\multirow{2}{*}{\rot{#1}}}
  \centering
  \begin{tabular}{lr|cc}
     & & Mean Acc.  $\uparrow$ & bits/MP $\uparrow$  \\ \toprule
     & Baluja~\cite{baluja2017hiding} & 0.51 & 0.5 \\
     & HiDDeN~\cite{zhu2018hidden} & 0.65 & 125 \\
     & LFM~\cite{wengrowski2019lfm} (printed) & 0.61 & 287 \\
     & LFM~\cite{wengrowski2019lfm} (screen) & 0.93 & 1109 \\ \hline
    % \multirow{4}{*}[1.5ex]{\rotcell{Ours}}&
    \multirow{4}{*}[-1.0ex]{\rot{Ours}}& 
        None & 0.49 & 0.1 \\
      & Pixelwise & 0.51 & 0.2 \\
      & Spatial & 0.89 & 318 \\
      & All & 0.99 & 571 \\[0.5ex]
  \end{tabular}
  \caption{Quantitative comparison of other methods and our ablations. We show numbers in terms of fraction of bits correctly recovered (mean accuracy) as well as bits-per-megapixel (bits/MP). Higher is better for both metrics. The bits/MP metric normalizes the message length and image sizes between different methods. All methods except ``LFM~\cite{wengrowski2019lfm} (screen)'' (cellphone camera/cellphone screen) are reported on the cellphone camera/consumer printer pipeline. We report LFM's results in this additional case because it was explicitly designed for screen/camera transmission.}
  \label{tab:bpp}
\end{table}

\begin{table}[t]
\centering
\begin{tabular}{l|ccc}
 & PSNR $\uparrow$     & SSIM $\uparrow$    & LPIPS $\downarrow$  \\
\hline
Baluja~\cite{baluja2017hiding} & 24.61  & 0.926  & 0.256 \\
HiDDeN~\cite{zhu2018hidden} (native) & 31.07  & 0.940  & 0.070 \\
HiDDeN~\cite{zhu2018hidden} & 24.55  & 0.775 & 0.202 \\
LFM~\cite{wengrowski2019lfm} & 20.89  & 0.910  & 0.315 \\
Ours & 27.25  & 0.927  & 0.194 \\
\end{tabular}
\caption{Quantitative comparison of encoded image quality, indicating how well hidden the message is. For HiDDeN~\cite{zhu2018hidden} we show both the metrics for the original lower resolution (native $128\times 128$) and upsampling to our compared resolution of $400\times 400$ with bicubic interpolation. At full resolution, our method produces an encoded image most similar to the original in all metrics.
}
\label{tab:supp_quality}
\end{table}

\section{Architecture Details}
Network architectures for our encoder (Table~\ref{tab:enc_net}) and decoder (Table~\ref{tab:dec_net}). Our detector uses the BiSeNet~\cite{BiSeNet} architecture.

\begin{table}[h]
\centering{
\resizebox{\linewidth}{!}{
\begin{tabular}{ccccccc}
\toprule
  \textbf{Layer} & \textbf{k} & \textbf{s} & \textbf{chns} & \textbf{in} & \textbf{out} & \textbf{input}
\tabularnewline
\midrule
inputs &  &  & 6   &  &  & image $+$ secret
\tabularnewline
conv1 & 3 & 1 & 6/32   & 1 & 1 & inputs
\tabularnewline
conv2 & 3 & 2 & 32/32   & 1 & 2 & conv1
\tabularnewline
conv3 & 3 & 2 & 32/64   & 2 & 4 & conv2
\tabularnewline
conv4 & 3 & 2 & 64/128   & 4 & 8 & conv3
\tabularnewline
conv5 & 3 & 2 & 128/256   & 8 & 16 & conv4
\tabularnewline
up6 & 2 & 1 & 256/128   & 16 & 8 & upsample(conv5)
\tabularnewline
conv6 & 3 & 1 & 256/128   & 8 & 8 & conv4 + up6
\tabularnewline
up7 & 2 & 1 & 128/64   & 8 & 4 & upsample(conv6)
\tabularnewline
conv7 & 3 & 1 & 128/64   & 4 & 4 & conv3 + up7
\tabularnewline
up8 & 2 & 1 & 64/32   & 4 & 2 & upsample(conv7)
\tabularnewline
conv8 & 3 & 1 & 64/32   & 2 & 2 & conv2 + up8
\tabularnewline
up9 & 2 & 1 & 32/32   & 2 & 1 & upsample(conv8)
\tabularnewline
conv9 & 3 & 1 & 70/32   & 1 & 1 & conv1 + up9 + inputs
\tabularnewline
conv10 & 3 & 1 & 32/32   & 1 & 1 & conv9
\tabularnewline
residual & 1 & 1 & 32/3   & 1 & 1 & conv10
\tabularnewline

\bottomrule
\end{tabular}}
}
\caption{Our encoder network architecture. \textbf{k} is the kernel size, \textbf{s} the stride, \textbf{chns} the number of input and output channels for each
layer, \textbf{in} and \textbf{out} are the accumulated stride for the input and output of each layer, \textbf{input} denotes the input of each layer with $+$ meaning concatenation and ``upsample'' performing $2\times$ nearest neighbor upsampling. A ReLU is applied after each layer except the last.}
\label{tab:enc_net}
\end{table}

\begin{table}[h]
\centering{
\resizebox{\linewidth}{!}{
\begin{tabular}{ccccccc}
\toprule
  \textbf{Layer} & \textbf{k} & \textbf{s} & \textbf{chns} & \textbf{in} & \textbf{out} & \textbf{input}
\tabularnewline
\midrule
conv1 & 3 & 2 & 3/32   & 1 & 2 & image
\tabularnewline
conv2 & 3 & 2 & 32/64   & 2 & 4 & conv1
\tabularnewline
conv3 & 3 & 2 & 64/128   & 4 & 8 & conv2
\tabularnewline
% \midrule
fc0   &   &   &    320000     &   &   & flatten(conv3)
\tabularnewline
fc1   &   &   &    320000/128     &   &   & fc0
\tabularnewline
fc2   &   &   &    128/6     &   &   & fc1
\tabularnewline
image\_warped   &   &   &    3/3     &   &   & transf(image, fc2)
\tabularnewline
\midrule

conv1 & 3 & 2 & 3/32   & 1 & 2 & image\_warped
\tabularnewline
conv2 & 3 & 1 & 32/32   & 2 & 2 & conv1
\tabularnewline
conv3 & 3 & 2 & 32/64   & 2 & 4 & conv2
\tabularnewline
conv4 & 3 & 1 & 64/64   & 4 & 4 & conv3
\tabularnewline
conv5 & 3 & 2 & 64/64   & 4 & 8 & conv4
\tabularnewline
conv6 & 3 & 2 & 64/128   & 8 & 16 & conv5
\tabularnewline
conv7 & 3 & 2 & 128/128   & 16 & 32 & conv6
\tabularnewline
fc0   &   &   &    20000     &   &   & flatten(conv7)
\tabularnewline
fc1   &   &   &    20000/512     &   &   & fc0
\tabularnewline
secret   &   &   &    512/100     &   &   & fc1
\tabularnewline

\bottomrule
\end{tabular}}
}
\caption{Our decoder network architecture. We indicate convolutional layers with the prefix ``conv'' and fully connected layers with the prefix ``fc.'' The first half of the network outputs an affine warp that is applied using a differentiable spatial transformer layer (``transf''). The warped result is fed into the second part of the network. A ReLU is applied after each layer except the last layer before the spatial transformer.}
\label{tab:dec_net}
\end{table}

\section{Code}
The code and pretrained networks can be found at \href{https://github.com/tancik/StegaStamp}{https://github.com/tancik/StegaStamp}.

\end{document}